\documentclass[nonacm, sigconf, screen]{acmart}
\renewcommand\footnotetextcopyrightpermission[1]{} 

\usepackage{multirow}
\usepackage{makecell}
\AtBeginDocument{%
  }

\setcopyright{acmlicensed}
\copyrightyear{2025}
\acmYear{2025}
\acmDOI{XXXXXXX.XXXXXXX}
\acmConference[ACM MM, 2025]{Make sure to enter the correct
  conference title from your rights confirmation email}{Oct 27--31, 2025}{Dublin, Ireland}

\acmSubmissionID{3796}

\begin{document}

\title[]{VC-LLM: Automated Advertisement Video Creation from Raw Footage using Multi-modal LLMs} 


\author{Dongjun Qian}
\affiliation{%
  \institution{Bytedance Inc}
  \city{}
  \country{}
}

\author{Kai Su}
\affiliation{%
  \institution{Bytedance Inc}
  \city{}
  \country{}
}

\author{Yiming Tan}
\affiliation{%
  \institution{Bytedance Inc}
  \city{}
  \country{}
}

\author{Qishuai Diao}
\affiliation{%
  \institution{Bytedance Inc}
  \city{}
  \country{}
}

\author{Xian Wu}
\affiliation{%
  \institution{Bytedance Inc}
  \city{}
  \country{}
}

\author{Chang Liu}
\affiliation{%
  \institution{Bytedance Inc}
  \city{}
  \country{}
}

\author{Bingyue Peng}
\affiliation{%
  \institution{Bytedance Inc}
  \city{}
  \country{}
}

\author{Zehuan Yuan}
\affiliation{%
  \institution{Bytedance Inc}
  \city{}
  \country{}
}









\begin{abstract}
As short videos have risen in popularity, the role of video content in advertising has become increasingly significant.
Typically, advertisers record a large amount of raw footage about the product and then create numerous different short-form advertisement videos based on this raw footage.
Creating such videos mainly involves editing raw footage and writing advertisement scripts, which requires a certain level of creative ability. It is usually challenging to create many different video contents for the same product, and manual efficiency is often low.
In this paper, we present VC-LLM, a framework powered by Large Language Models for the automatic creation of high-quality short-form advertisement videos. Our approach leverages high-resolution spatial input and low-resolution temporal input to represent video clips more effectively, capturing both fine-grained visual details and broader temporal dynamics. In addition, during training, we incorporate supplementary information generated by rewriting the ground truth text, ensuring that all key output information can be directly traced back to the input, thereby reducing model hallucinations. We also designed a benchmark to evaluate the quality of the created videos. Experiments show that VC-LLM based on GPT-4o can produce videos comparable to those created by humans. Furthermore, we collected numerous high-quality short advertisement videos to create a pre-training dataset and manually cleaned a portion of the data to construct a high-quality fine-tuning dataset. Experiments indicate that, on the benchmark, the VC-LLM based on fine-tuned LLM can produce videos with superior narrative logic compared to those created by the VC-LLM based on GPT-4o.
\end{abstract}

\ccsdesc[500]{Information systems~Multimedia information systems}

\keywords{
Multimodal AI, Large Language Models, Advertisement video creation, Script generation}


\maketitle
\renewcommand{\shortauthors}{} 

\begin{figure*}[t]
    \centering
    \includegraphics[width=0.85\linewidth]{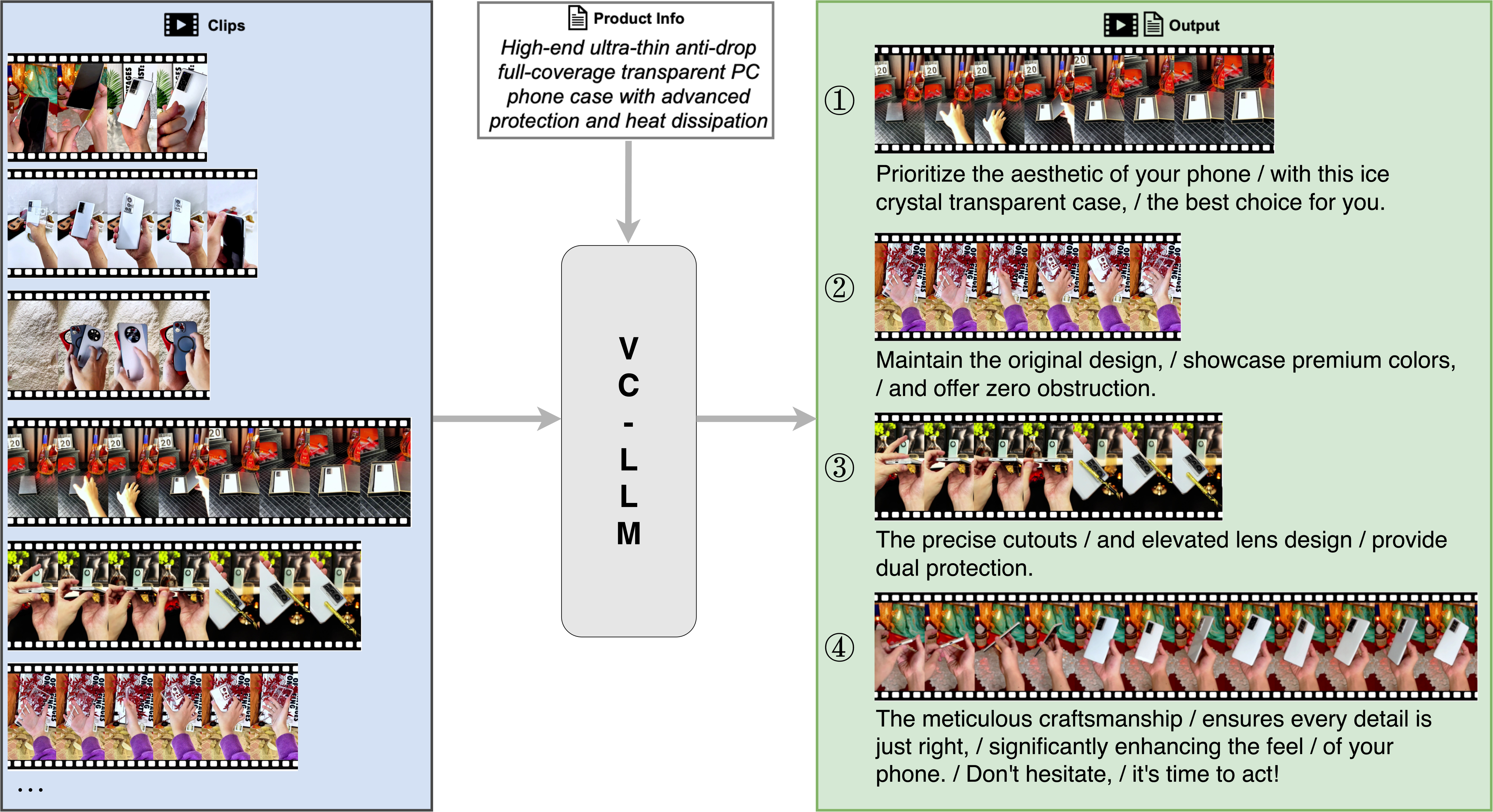}
    \caption{Overview of the VC-LLM Framework: Input product information and video clips, output a sequence of selected video clips, corresponding scripts, and segmented subtitles.
    }
    \label{fig:framework}
\end{figure*}

\section{Introduction}
In recent years, short-form video content has surged in popularity, reshaping the landscape of advertising and digital marketing\citep{ge2021effect}. 
Advertisers increasingly rely on concise, visually engaging videos to promote products and services. However, producing a diverse set of high-quality advertisement videos remains a labor-intensive process. Typically, advertisers gather extensive raw footage and then manually edit this material, while concurrently developing tailored advertisement scripts—a process that demands significant creative input and specialized editing skills. This traditional workflow often suffers from low efficiency, particularly when multiple variations are required to target different market segments.

The advent of Large Language Models (LLMs)\citep{zhao2023survey,minaee2024large} and their multi-modal counterparts\citep{yin2023survey,jin2024efficient} has introduced new possibilities for automating content creation. Recent advancements in LLMs have demonstrated their ability to perform complex tasks such as natural language understanding and generation, while multi-modal LLMs extend these capabilities by processing both text and visual information simultaneously. Despite this progress, current applications of LLMs in video production have largely focused on tasks such as caption generation or text-based video summarization\citep{wang2024lave}, leaving the domain of full-scale advertisement video creation relatively underexplored.

To bridge this gap, we propose VC-LLM, a novel framework that leverages the power of LLMs to automatically create high-quality short-form advertisement videos from raw footage. VC-LLM not only streamlines the video editing process but also integrates an automated script generation mechanism, which, in tandem with advanced subtitle segmentation, ensures that the final videos exhibit coherent narrative logic and visual appeal. This framework is built upon state-of-the-art models such as GPT-4o\citep{hurst2024gpt} and further refined through a dedicated fine-tuning process on open-source models\citep{internlmxcomposer2_5,internlmxcomposer} using a high-quality dataset curated from existing short advertisement videos.

Our work contributes in several key areas. 
First, we adopt a dual-resolution encoding strategy, using high-resolution spatial inputs and low-resolution temporal inputs to more effectively represent video clips.
Second, we incorporate supplementary information derived from ground-truth scripts during training. This ensures that all key information in the output is grounded in the input, thereby mitigating hallucination and improving factuality.
Third, we design a comprehensive benchmark to evaluate the quality of created videos, considering multiple aspects, including the alignment between visual content and oral script, narrative logic of the visual content, factuality, contextual coherence, logical correctness of the script, word count discrepancy between the script and the target value suited to the corresponding visual content, and subtitle segmentation accuracy. 
Fourth, our experiments indicate that VC-LLM, when powered by GPT-4o, can produce advertisement videos that are better than human-created content in terms of alignment between visual content and oral script, as well as script contextual coherence and logical correctness. More importantly, the fine-tuned version of our framework demonstrates superior narrative logic, suggesting that targeted fine-tuning on curated data can significantly enhance the creative outputs of automated video creation systems.


By automating the traditionally manual process of advertisement video creation, VC-LLM has the potential to reduce production costs and accelerate content deployment, enabling advertisers to rapidly create multiple tailored videos for different target demographics. The proposed framework thus represents a significant step toward the democratization of video production, opening new avenues for personalized advertising at scale.

In the following sections, we detail the architecture of VC-LLM, describe the methodology for constructing our pre-training and fine-tuning datasets, and present our experimental results on the proposed benchmark. Through this work, we aim to contribute to the growing body of research on multi-modal content generation and offer practical insights for the future of automated advertising.
\begin{figure*}[t]
    \centering
    \includegraphics[width=0.9\linewidth]{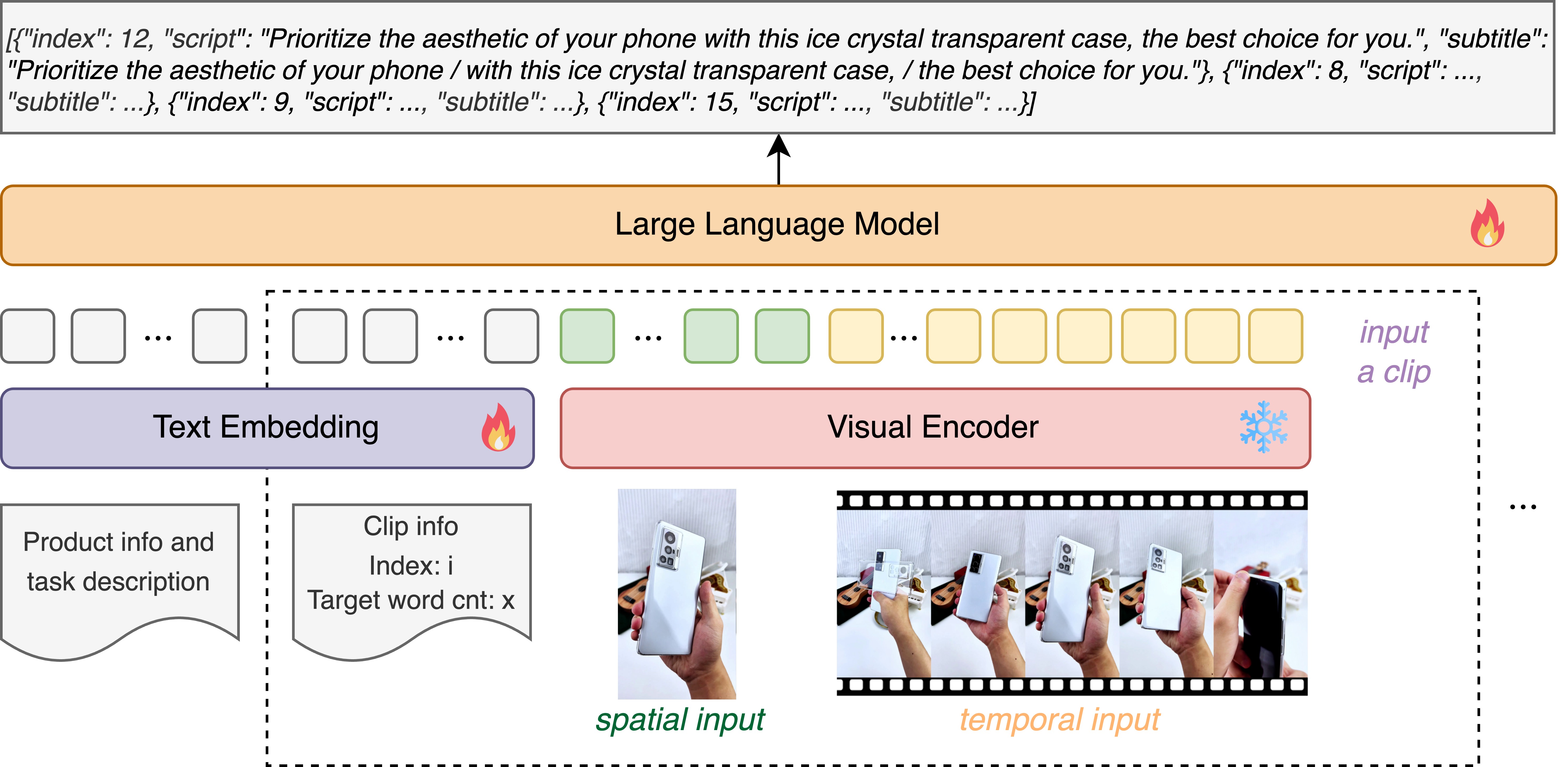}
    \caption{The architecture of the model. 
    The model accepts product information and an indexed series of video clips as input, and it generates an output sequence in which each element consists of a selected video clip, its associated script and subtitle segmentation.
    During training, the parameters of the language model and the text embedding module are updated, whereas the visual encoder’s parameters remain fixed.
    Text content is processed via the text embedding module to yield an embedding sequence derived from tokenization, while visual content is transformed by the visual encoder into its corresponding embedding sequence.
    Green blocks denote tokens associated with the spatial input, and yellow blocks denote tokens corresponding to the temporal input.
    }
    \label{fig:model}
\end{figure*}

\section{Related Work}
Recent advancements in Large Language Models (LLMs)\citep{zhao2023survey,minaee2024large,ouyang2022training,grattafiori2024llama,liu2024deepseek,cai2024internlm2,qwen2.5} have enabled these models to perform increasingly complex tasks. Furthermore, Multi-modal Large Language Models (MLLMs)\citep{yin2023survey,jin2024efficient,internlmxcomposer2_5,glm2024chatglm,chen2024internvl,Qwen2.5-VL,reid2024gemini,hurst2024gpt} extend LLMs' capabilities to process both visual and textual content directly. SmartEdit~\citep{huang2024smartedit} integrates MLLMs with a bidirectional interaction module and diffusion models, enhancing the latter’s reasoning and understanding for image editing. Similarly, GILL~\citep{koh2024generating} advances the integration of MLLMs with text-to-image generation models, facilitating the coherent generation of images and text based on both modalities. GG-Editor~\citep{xu2024gg} introduces a GPT-guided local avatar editing framework, leveraging MLLMs to predict or select specific regions for modification. Despite these advancements, little attention has been given to the automated creation of advertisement videos from raw footage. This process involves interleaved tasks, including video clip selection, visually grounded oral script generation, and subtitle segmentation.
To address this gap, this work proposes a framework for the automated creation of advertisement videos.

\section{Method}
\subsection{Framework}
As illustrated in Fig.~\ref{fig:framework},
the proposed framework takes product information and numerous material video clips as input, and outputs selected video clips along with corresponding scripts and subtitle segmentation.
Product information primarily includes the product name, selling points, and other relevant details.
$P$ is used to represent the product information.
The input video clips are short segments extracted from raw footage.
We use $C=\{\text{clip}_1,\ldots,\text{clip}_N\}$ to represent the input video clips, where $N$ is the number of the clips. 
Thus the framework can be formulated as
\begin{equation}
Y=f(P,C,K)
\end{equation}
where $f(\cdot)$ is a function based on MLLMs. $K$ represents the desired number of clips to be selected, but $K$ is optional. 
$Y$ is the output sequence where each element $y_i\in Y$ is a tuple containing a selected video clip, its script, and subtitle segmentation:
\begin{equation}
y_i=(\text{clip}_{j_i},\text{script}_{j_i},\text{subtitles}_{j_i}), j_i\in\{1,\ldots,N\}
\end{equation}
represents the output, which can be parsed into a protocol that can be used to produce a video.


\subsection{Model}
As illustrated in Fig.~\ref{fig:model}, 
the model comprises three main components: a visual encoder, a text embedding module, and a large language model.
Text content is transformed via the text embedding module:
\begin{equation}
\text{TextEmb}: \text{Text}\rightarrow \mathbb{R}^d
\end{equation}
Visual content is encoded by the visual encoder:
\begin{equation}
\text{VisEnc}: \text{Visual}\rightarrow \mathbb{R}^d
\end{equation}
where $d$ denotes the embedding dimension of the large language model.
The visual and textual tokens are then merged and fed into the large language model for processing, generating the output $Y$.
During training, let $\theta_{LLM}$ and $\theta_{text}$ denote the parameters of the language model and text embedding, respectively, and $\theta_{visual}$ denote the parameters of the visual encoder. The updates are:
\begin{equation}
\theta_{LLM},\theta_{text}\leftarrow \text{Optimizer}(\theta_{LLM},\theta_{text},\nabla E)
\end{equation}
where $\nabla E$ is the gradient of the loss function with respect to the model parameters. $\theta_{visual}$ remains fixed.

\subsection{Spatial and Temporal Clip Representation}
\begin{figure}[htbp]
    \centering
    \includegraphics[width=1.0\linewidth]{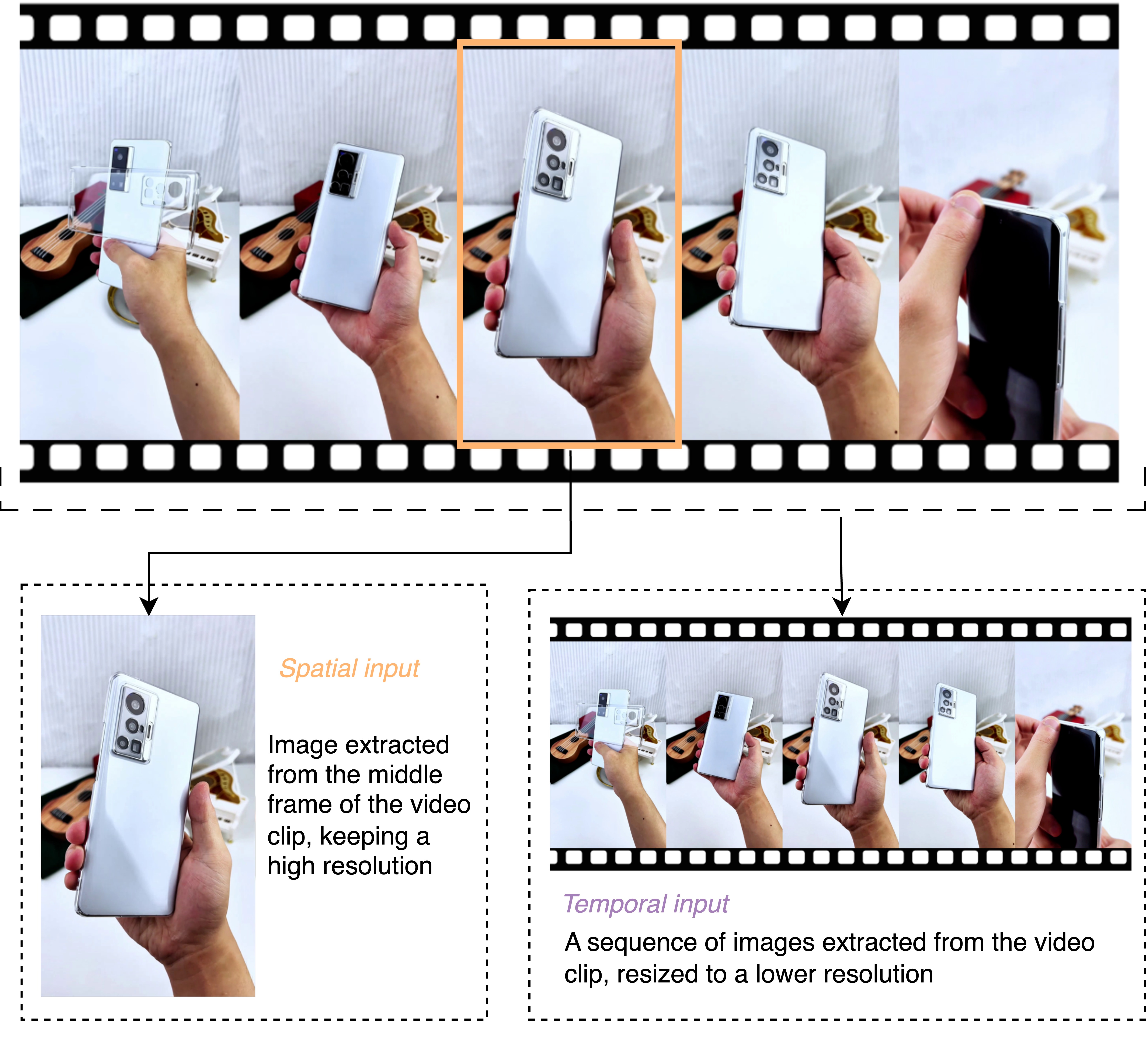}
    \caption{The spatial and temporal representation of a video clip. 
    The spatial input is a higher resolution image of the middle frame, providing detailed information about the product and environment. The temporal input is a sequence of images covering the entire video clip and providing motion information.
    }
    \label{fig:spatial_temporal}
\end{figure}

Due to the need to input numerous video clips, in order to control the length of the input tokens while effectively representing the video clips, we propose using spatial and temporal inputs to jointly represent the video clips.
Use V to represent the visual information of a video clip, as described by the following formula.
\begin{equation}
V=\{v_{t} | t\in [0,T]\}
\end{equation}
where $T$ is the duration of the video clip.
Then, the spatial input can be described by the following expression.
\begin{equation}
X_{spatial}=resize(v_{t},r_{s}),t=\left\lfloor T/2\right\rfloor
\end{equation}
where $resize(\cdot,r)$ denotes the function that resizes images to resolution $r$, and $r_{spatial}$ represents the resolution of the spatial input.
The temporal input can be described by the following expression.
\begin{equation}
X_{temporal}= 
\begin{cases}
    resize(uniform(M,l), r_{temporal}) & \text{if } l < m+1\\
    resize(M, r_{temporal}) & \text{if } l \geq m+1
\end{cases}
\end{equation}
\begin{equation}
M=\{v_0,v_h,\ldots,v_{m\cdot h}\},m=\left\lfloor T/h\right\rfloor
\end{equation}
where $h=1/\text{fps}$ represents the frame interval, $l$ represents the maximum number of frames, and $r_{temporal}$ represents the resolution of the temporal input.
As also shown in Fig.~\ref{fig:spatial_temporal}, the spatial input is a higher resolution image of the middle frame, primarily used to describe detailed information of the product and the surrounding environment.
The temporal input consists of a sequence of lower resolution images that cover the entire video clip, used to describe the events occurring throughout the clip and actions within the clip.

\subsection{Reduction of Model Hallucination}
\begin{figure}[H]
    \centering
    \includegraphics[width=1.0\linewidth]{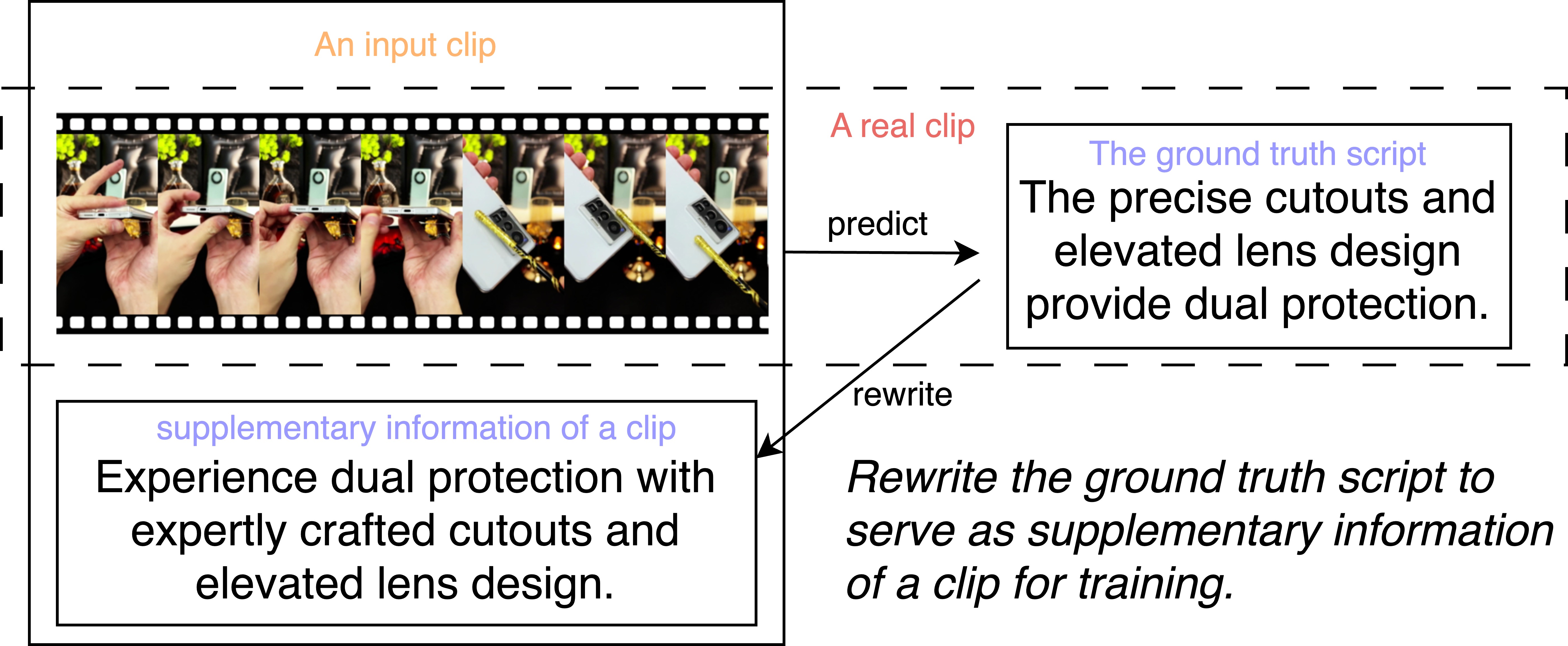}
    \caption{Clip representation with extra information. During SFT, we use the rewritten script as additional input information to ensure that the critical information required for predicting the ground truth script can be found in the input, thereby reducing model hallucination. }
    \label{fig:prompt_script}
\end{figure}

We address model hallucination by focusing on improvements in training data.
For a given training task, if some key information in the ground truth output is either absent from the input data or present in the visual data but not effectively extracted or understood, the model is forced to generate unsupported content, thereby exacerbating the phenomenon of model hallucination.
Therefore, as illustrated in Fig.~\ref{fig:prompt_script}, in the SFT data, we rewrite the ground truth script and incorporate it as additional supplementary information for the video clips in the input, aiming to mitigate hallucination.
Let $X$ denote the original input, $Z$ denote the supplementary information obtained by rewriting the ground truth script, 
$\tilde{X}=X\bigoplus Z$ represent the augmented input, $Y$ denote the ground truth output, and $f_{\theta}(\cdot)$ be the model parameterized by $\theta$, then the training objective can be defined as:
\begin{equation}
\min_{\theta}\mathcal{L}(f_{\theta}(\tilde{X}),Y)
\end{equation}
where $\mathcal{L}$ is the loss function measuring the discrepancy between the model output and the ground truth.

\begin{figure*}[t]
    \centering
    \includegraphics[width=1.0\linewidth]{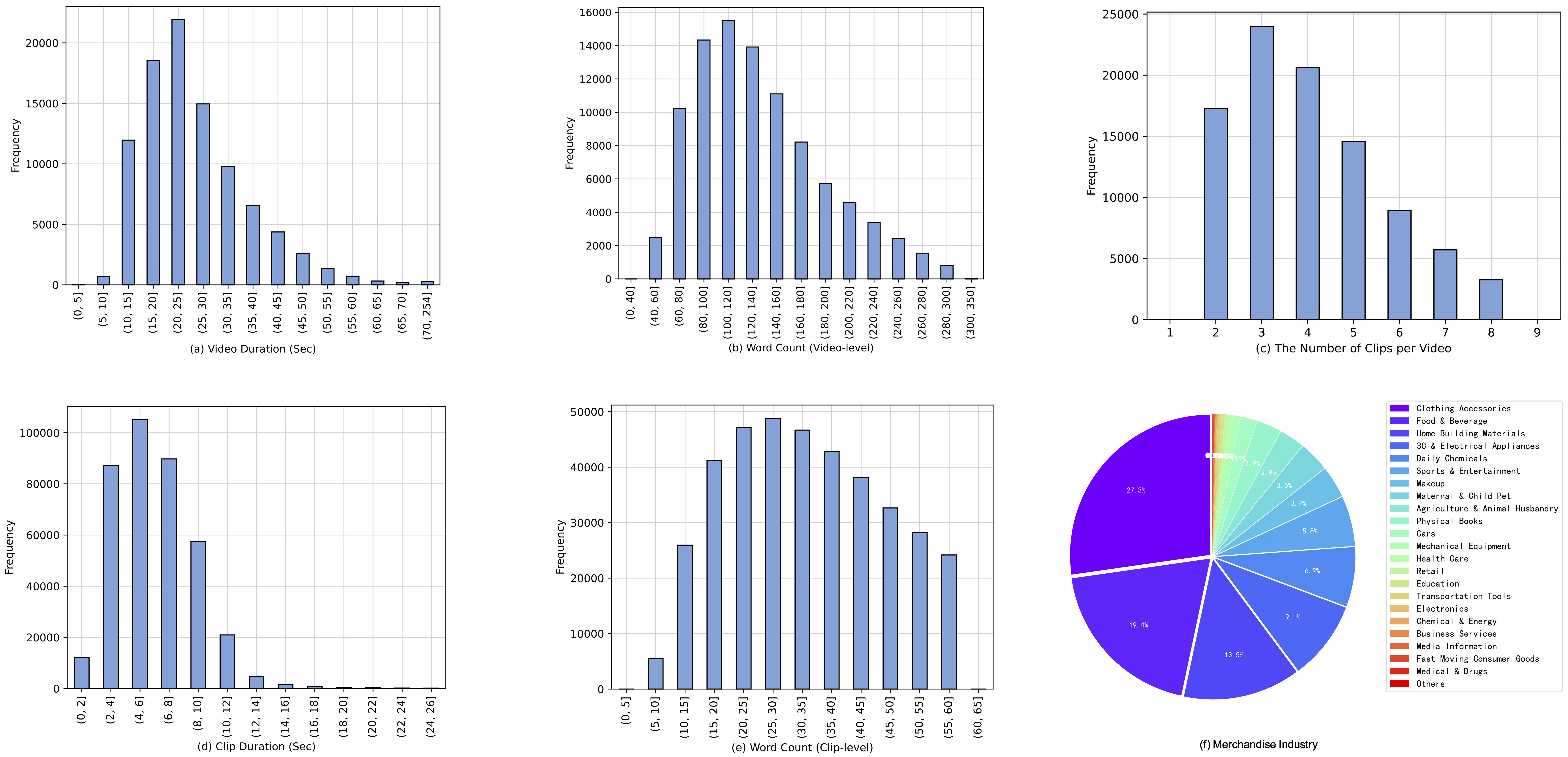}
    \caption{
    Statistics of the dataset: (a) Video Duration - The distribution of overall video lengths. (b) Word Count (Video-level) - The distribution of overall word counts across videos. (c) Number of Clips per Video - The distribution of the typical number of separate clips that are taken from each video. (d) Clip Duration - The distribution of overall video clip lengths. (e) Word Count (Clip-level) - The distribution of overall word counts across individual video clips. (f) Merchandise Industry - The distribution of merchandise industries.}
    \label{fig:stat}
\end{figure*}
\begin{figure}[t]
    \centering
    \includegraphics[width=1.0\linewidth]{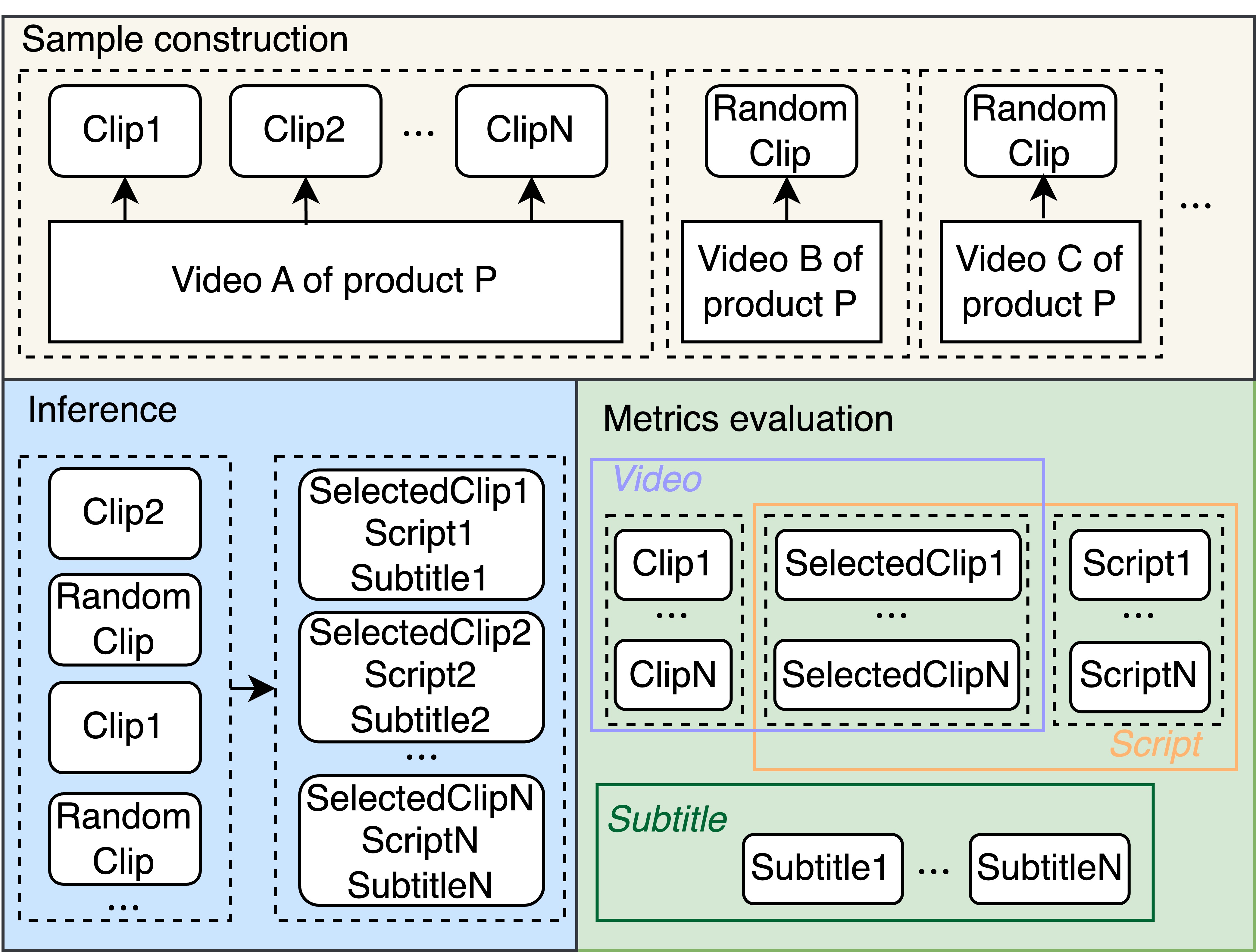}
    \caption{Illustration of the benchmark.
    Sample Construction: Each sample consists of all video clips from its original video, along with additional clips extracted from other videos featuring the same product.
    Inference: Based on all video clips included in the sample, VC-LLM is used to create the final video.
    Metrics Evaluation: 
    Then, the created video, i.e., the selected segment sequence, generated script, and subtitles will be evaluated based on defined metrics.
    }
    \label{fig:benchmark}
\end{figure}

\section{Datasets}

\subsection{Dataset construction}
Our dataset construction process consists of four main stages: data selection, data parsing, data processing, and data splitting.

\noindent\textbf{Data Selection.}
We begin by collecting publicly available advertisement videos from domestic platforms. In total, we collect $1.57$ million video contents corresponding to $389,000$ unique products.

\noindent\textbf{Data Parsing.} 
Each video content is segmented into a sequence of video clips. For videos that include oral scripts, we employ an Automatic Speech Recognition (ASR) model to extract both the scripts and their corresponding timestamps, and subsequently segment the video based on punctuation-aligned timestamps. For videos lacking oral scripts, we measure the visual feature differences between consecutive frames and segment the video at points where these differences exceed a predefined threshold.

\noindent\textbf{Data Processing.}
This stage focuses on filtering high-quality video contents and refining oral scripts through human annotation. 
The filtering criteria require that (1) the video duration does not exceed $120$ seconds, (2) the number of video clips falls between $2$ and $8$, (3) the oral scripts are fluent and free of typographical errors, and (4) the scripts are semantically relevant to both the video content and the product information. The evaluation of criteria (3) and (4) is conducted using GPT-4o. After this stage, $230,000$ videos corresponding to $99,000$ unique products remain. Additionally, in the supervised fine-tuning subset, oral scripts are manually refined and segmented following the data splitting process.

\noindent\textbf{Data Splitting.}
The dataset is partitioned to facilitate both training and evaluation. First, $5,000$ distinct products are sampled, with one video per product reserved for the test set. The remaining products ($94,000$) contribute to the training set, with at most two video creatives retained per product, yielding a training set of $110,000$ videos. The training set is further subdivided into a $100,000$-video continued pre-training set and a $10,000$-video supervised fine-tuning set. 
Notably, all training videos contain oral scripts and are segmented using ASR. 
For the test set, we additionally collected random video clips, including those without oral scripts, from other videos of the same product to enhance diversity and robustness in evaluation. 
The test set is used to construct the benchmark. Details are provided in the Benchmark section.

\noindent\textbf{Data Statistics.}
Fig.~\ref{fig:stat} presents the statistical overview of our final dataset. At the video level, durations primarily range from $15$ to $30$ seconds, with the number of Chinese characters per video typically falling between $80$ and $140$. Most videos consist of $2$ to $4$ clips. At the clip level, durations generally range from $2$ to $8$ seconds, with character counts between $20$ and $35$. Regarding industry distribution, advertisements for clothing accessories, food and beverages, and home building materials represent the largest shares of the dataset.

\subsection{Task Construction}
Based on the constructed dataset, we define four tailored training tasks—three fundamental tasks and one compound task—with only the compound task used during testing. The fundamental tasks target individual modalities: for the video track, the remix task takes as input the product information along with an unordered set of video clips, which includes all clips from the corresponding video plus several random clips from other videos, and requires the model to reconstruct the correct ordered sequence of clips. For the audio track, the script predicting task provides the product information and the ordered sequence of video clips, and its objective is to predict the corresponding oral scripts extracted during data processing. And for the subtitle track, the script segmentation task presents the product information, the ordered clips, and the extracted oral scripts, with the goal of punctuating these scripts accurately, as validated by human annotation. The compound task integrates all three modalities, requiring the model to select and sequence video clips, generate corresponding oral scripts, and segment the scripts appropriately. In this task, the input consists of the product information and an unordered set of clips drawn from both the target video and other random videos, while the ground truth comprises a sequence of tuples, each containing a selected video clip, its oral script, and the corresponding segmented subtitles.





\begin{table*}[t]
\centering
\caption{Metrics evaluated on the benchmark}
\label{tab:metrics}{
\begin{tabular}{ccccc|c|ccccc|c}
\toprule
\multirow{2}{*}{Model} 
& \multirow{2}{*}{\makecell{Input supplementary\\info during SFT}} 
& \multirow{2}{*}{\makecell{Max\\frames}}
& \multirow{2}{*}{\makecell{Continued\\pre-training}} 
& \multirow{2}{*}{\makecell{Spatial\\input}}
& \multicolumn{1}{c|}{Video}
& \multicolumn{5}{c|}{Script}  
& \multicolumn{1}{c}{Subtitle} \\

&&&&
& $SRA\uparrow$ 
& $VSC\uparrow$ 
& $Fact\uparrow$ 
& $Coh\uparrow$ %
& $Logic\uparrow$
& $WCD\downarrow$ %
& $SSA\uparrow$ 
\\ 
\midrule

Human
& -
& -
& -
& -
& -
& $1.6276$
& -
& $1.5590$
& $1.8663$
& -
& -
\\

GPT-4o 
& -
& 5
& -
& -
& $0.0281$
& $\textbf{1.8532}$ 
& $1.9532$ 
& $1.9283$ 
& $1.9820$  
& $7.2442$ 
& $0.3334$ 
\\

IXC1.0 
& -
& 1
& -
& -
& $0.0424$ 
& $1.6918$ 
& $1.8125$ 
& $1.8949$ 
& $1.9467$ 
& $2.1961$ 
& $0.9030$ 
\\

IXC2.5
& -
& $1$
& -
& -
& $0.0382$ 
& $1.7293$ 
& $1.8395$ 
& $1.9112$ 
& $1.9623$ 
& $2.2850$ 
& $\textbf{0.9268}$ 
\\

IXC2.5
& $\checkmark$
& $1$
& -
& -
& $0.0414$ 
& $1.7991$ 
& $1.9427$ 
& $\textbf{1.9435}$ 
& $\textbf{1.9870}$ 
& $5.3476$ 
& $0.8896$ 
\\

IXC2.5
& $\checkmark$
& $5$
& -
& -
& $0.0521$ 
& $1.8018$ 
& $1.9474$ 
& $1.9387$ 
& $1.9799$ 
& $4.5411$ 
& $0.9076$ 
\\

IXC2.5
& $\checkmark$
& 5
& $\checkmark$
& -
& $0.1076$ 
& $1.8055$ 
& $1.9573$ 
& $1.9431$ 
& $1.9826$ 
& $1.6838$ 
& $0.9114$ 
\\

IXC2.5
& $\checkmark$
& 5
& $\checkmark$
& $\checkmark$
& $\textbf{0.1098}$ 
& $1.8069$ 
& $\textbf{1.9684}$ 
& $1.9408$ 
& $1.9752$ 
& $\textbf{1.1285}$ 
& $0.9031$ 
\\

\bottomrule 
\end{tabular}
}
\end{table*}
\section{Benchmark}

\subsection{Benchmark Construction}
As illustrated in Fig.~\ref{fig:benchmark}, 
each sample includes all video clips from its original source, supplemented with additional clips sourced from other videos featuring the same product. Using all video clips within the sample, VC-LLM is utilized to create the final video. Subsequently, the created video, comprising the selected segment sequence, generated script, and subtitles, is evaluated against predefined metrics.

\subsection{Evaluation Metrics}
We define six metrics, described as follows.

\noindent\textbf{SRA} (selection and rank accuracy) assesses the narrative logic.
The input sequence of clips is represented by $C=\{clip_1,\ldots,clip_N\}$.
The ground truth sequence of clips is a subset of $C$ and is represented by $G=\{g_1,\ldots,g_K\}$, where $K\leq N$ and the ordering reflects the correct sequence of clips. The predicted sequence of video clips is also a subset of $C$ and is represented by $S=\{s_1,\ldots,s_K\}$,
with the ordering of elements corresponding to the predicted selection and arrangement. Then, the selection and rank accuracy is defined as
\begin{equation}
SRA=\prod_{i}^{K}1(g_i=s_i)
\end{equation}
where $1(\cdot)$ is an indicator function that returns $1$ if the predicted clip $s_i$ matches the ground truth clip $g_i$ at the $i$-th position, and $0$ otherwise.

\noindent\textbf{VSC} (visual script correlation) 
evaluates the semantic alignment between the selected video clips and the generated scripts. The VSC metric is assigned a value from $\{0,1,2\}$, with the evaluation conducted using GPT-4o.

\noindent\textbf{Fact} (factulity) 
assesses whether all key information in the generated script is supported by the provided product information and the selected video clips. The Fact metric is assigned a value from $\{0,1,2\}$, with the evaluation performed using GPT-4o.

\noindent\textbf{Coh} 
measures the contextual coherence of the script sequence. The Coh metric is assigned a value from $\{0,1,2\}$, with the assessment conducted using GPT-4o.

\noindent\textbf{Logic} 
evaluates the logical correctness of the script. The Logic metric is assigned a value from $\{0,1,2\}$, with the assessment conducted using GPT-4o.

\noindent\textbf{WCD} (word count discrepancy) 
quantifies the deviation of the script's word count from the expected length based on the corresponding video clip. It is computed as:
\begin{equation} 
WCD=|WordCount_{script} - WordCount_{target}|
\end{equation}

\noindent\textbf{SSA} (subtitle segmentation accuracy) 
measures the accuracy of subtitle segmentation based on the following criteria:
\begin{enumerate}
    \item The character count of each segment must not exceed a predefined limit.
    \item Text between two punctuation marks should not be split into segments that exceed a predefined multiple of the character count.
    \item Essential words should not be divided across different segments.
\end{enumerate}
If all criteria are met, the SSA score is set to $1$; otherwise, it is $0$. The SSA metric serves as an indicator of the readability of the generated subtitles.

\begin{table*}[htbp]
\centering
\caption{Impact of Spatial Input}
\label{tab:spatial_input}
\begin{tabular}{c|ccccccc}
\toprule
Spatial Input
&
$SRA\uparrow$ 
& $VSC\uparrow$ 
& $Fact\uparrow$ 
& $Coh\uparrow$ 
& $Logic\uparrow$ 
& $WCD\downarrow$ 
& $SSA\uparrow$ \\ 
\midrule

-  
& $0.1076$  
& $1.8055$  
& $1.9573$  
& $\textbf{1.9431}$  
& $\textbf{1.9826}$  
& $1.6838$  
& $\textbf{0.9114}$ \\ 

$\checkmark$  
& $\textbf{0.1098}$  
& $\textbf{1.8069}$  
& $\textbf{1.9684}$  
& $1.9408$  
& $1.9752$  
& $\textbf{1.1285}$  
& $0.9031$ \\ 

\bottomrule
\end{tabular}
\end{table*}
\begin{table*}[htbp]
\centering
\caption{Impact of Input Supplementary Info During SFT}
\label{tab:condition}
\begin{tabular}{c|ccccccc}
\toprule
\makecell{Input supplementary\\info during SFT}
&
$SRA\uparrow$ 
& $VSC\uparrow$ 
& $Fact\uparrow$ 
& $Coh\uparrow$ 
& $Logic\uparrow$ 
& $WCD\downarrow$ 
& $SSA\uparrow$ \\ 
\midrule

-  
& $0.0382$  
& $1.7293$  
& $1.8395$  
& $1.9112$  
& $1.9623$  
& $\textbf{2.2850}$  
& $\textbf{0.9268}$ \\ 

$\checkmark$  
& $\textbf{0.0414}$  
& $\textbf{1.7991}$  
& $\textbf{1.9427}$  
& $\textbf{1.9435}$  
& $\textbf{1.9870}$  
& $5.3476$  
& $0.8896$ \\ 

\bottomrule
\end{tabular}
\end{table*}
\begin{table*}[htbp]
\centering
\caption{Impact of Continued Pre-training}
\label{tab:continued_pretraining}
\begin{tabular}{c|ccccccc}
\toprule
\makecell{Continued\\Pre-training}
&
$SRA\uparrow$ 
& $VSC\uparrow$ 
& $Fact\uparrow$ 
& $Coh\uparrow$ 
& $Logic\uparrow$ 
& $WCD\downarrow$ 
& $SSA\uparrow$ \\ 
\midrule

-  
& $0.0521$  
& $1.8018$  
& $1.9474$  
& $1.9387$  
& $1.9799$  
& $4.5411$  
& $0.9076$ \\ 

$\checkmark$  
& $\textbf{0.1076}$  
& $\textbf{1.8055}$  
& $\textbf{1.9573}$  
& $\textbf{1.9431}$  
& $\textbf{1.9826}$  
& $\textbf{1.6838}$  
& $\textbf{0.9114}$ \\ 

\bottomrule
\end{tabular}
\end{table*}
\begin{table*}[htbp]
\centering
\caption{Impact of Max Frames}
\label{tab:max_frames}
\begin{tabular}{c|ccccccc}
\toprule
Max frames
&
$SRA\uparrow$ 
& $VSC\uparrow$ 
& $Fact\uparrow$ 
& $Coh\uparrow$ 
& $Logic\uparrow$ 
& $WCD\downarrow$ 
& $SSA\uparrow$ \\ 
\midrule

1  
& $0.0414$  
& $1.7991$  
& $1.9427$  
& $\textbf{1.9435}$  
& $\textbf{1.9870}$  
& $5.3476$  
& $0.8896$ \\ 

5  
& $\textbf{0.0521}$  
& $\textbf{1.8018}$  
& $\textbf{1.9474}$  
& $1.9387$  
& $1.9799$  
& $\textbf{4.5411}$  
& $\textbf{0.9076}$ \\ 

\bottomrule
\end{tabular}
\end{table*}
\begin{figure*}[t]
    \centering
    \includegraphics[width=0.87\linewidth]{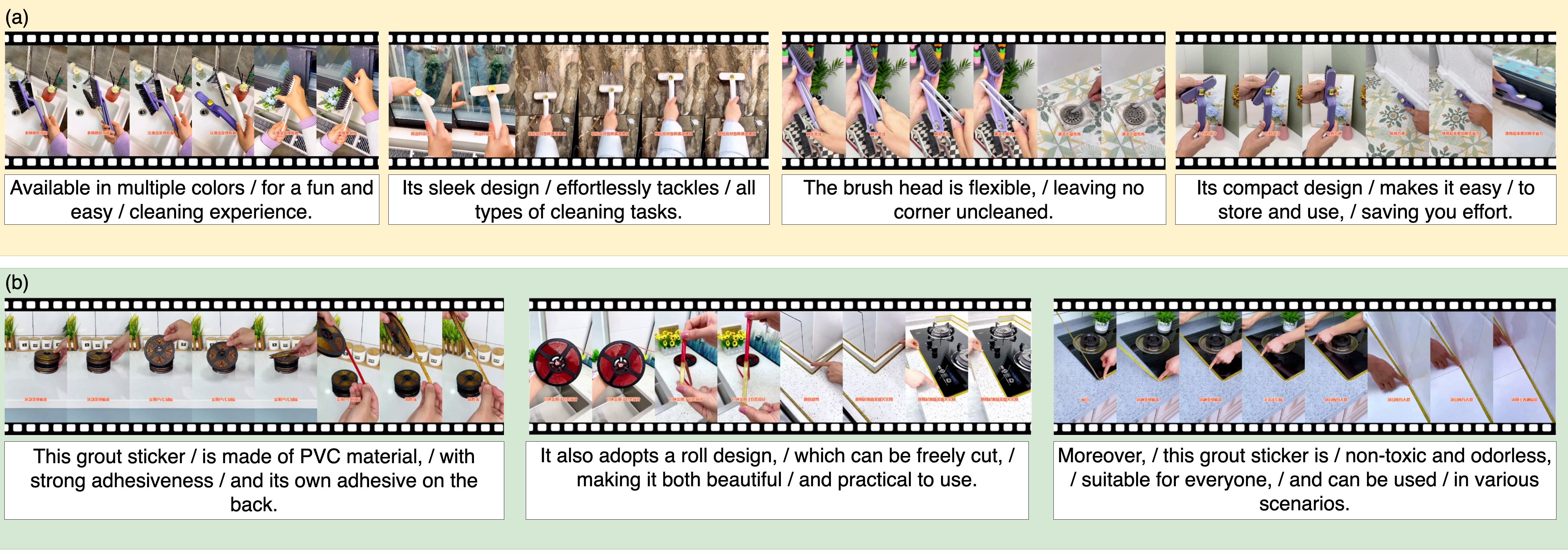}
    \caption{Videos created by VC-LLM.}
    \label{fig:showcases}
\end{figure*}

\begin{table*}[h]
\centering
\caption{Impact of Base Model}
\label{tab:model_comparison}
\begin{tabular}{c|ccccccc}
\toprule
Model
&
$SRA\uparrow$ 
& $VSC\uparrow$ 
& $Fact\uparrow$ 
& $Coh\uparrow$ 
& $Logic\uparrow$ 
& $WCD\downarrow$ 
& $SSA\uparrow$ \\ 
\midrule

IXC1.0  
& $\textbf{0.0424}$  
& $1.6918$  
& $1.8125$  
& $1.8949$  
& $1.9467$  
& $\textbf{2.1961}$  
& $0.9030$ \\ 

IXC2.5  
& $0.0382$  
& $\textbf{1.7293}$  
& $\textbf{1.8395}$  
& $\textbf{1.9112}$  
& $\textbf{1.9623}$  
& $2.2850$  
& $\textbf{0.9268}$ \\

\bottomrule
\end{tabular}
\end{table*}
\section{Experiments}

\subsection{Training Details}
We train our model using the open-source InternLM-XComposer (IXC) series\citep{internlmxcomposer,internlmxcomposer2_5} as the base architecture. A batch size of 1024 is used for continued pre-training, while a batch size of 256 is employed for supervised fine-tuning. 
During training, we update all parameters except those of the ViT module (OpenAI ViT-L/14\citep{radford2021learning} in IXC2.5 and EVA ViT-G/14\citep{fang2023eva} in IXC1.0). The updated parameters include the LLM model (InternLM series\citep{team2023internlm,cai2024internlm2}), the multilayer perceptron that projects visual features to the LLM’s feature dimension, and the BERT\citep{devlin2019bert}-based Q-Former\citep{li2022blip,li2023blip} used in IXC1.0.
The optimizer is AdamW\citep{loshchilov2017decoupled}, with a learning rate of $2\times10^{-5}$.
For each video clip, images are extracted at a rate of one frame per second, and up to five images are uniformly sampled from the resulting sequence. The spatial input images have a resolution of $996$ (height) $\times$ $560$ (width), whereas the temporal input frames are resized to $560$ (height) $\times$ $315$ (width). During continued pre-training, each sample is augmented by generating remix tasks four times, with the input clips arranged in different orders for each iteration, and an additional script prediction task is incorporated. Similarly, for supervised fine-tuning, each sample is used to generate compound tasks four times, with varied clip orders across iterations.

\subsection{Evaluation Details}
For inference, our model employs greedy search. The evaluation of metrics is divided between local Python scripts and GPT-4o. Specifically, the visual script correlation (VSC), factuality (Fact), contextual coherence (Coh), and logical correctness (Logic) metrics are assessed using GPT-4o. When conducting evaluations with GPT-4o, videos are sampled at one frame per second without an upper limit on frame count, and each created video is represented as a sequence of tuple of (video clip image sequence, script, subtitles). GPT-4o outputs values for VSC, Fact, Coh, Logic, and other relevant metrics. The selection and rank accuracy (SRA), word count discrepancy (WCD), and subtitle segmentation accuracy (SSA) metrics are computed using our local Python scripts. For SSA, a segment is considered incorrect if it contains more than $13$ units (with a Chinese character counted as 1 unit, an English letter as $0.4$ units, and a space as $0.5$ units). In addition, the maximum number of segments between two punctuation marks is set to $\left\lceil units/10 \right\rceil$. Any instance exceeding this limit indicates incorrect segmentation. Finally, Jieba\citep{jieba} Chinese text segmentation is employed to extract all essential words, and if an essential word is divided across segments, it is deemed incorrectly segmented.

\subsection{Experimental Results}
We evaluate the proposed metrics of the VC-LLM framework based on different LLM models on the benchmark, as shown in Table\ref{tab:metrics}. “Human” refers to evaluations of the VSC, Coh, and Logic metrics using the original ASR output. Since the ordering of segmented video clips in the sample is used as the ground truth for the SRA metric, the SRA is not assessed for human outputs. Likewise, because the human-provided scripts are assumed to be accurate, the Fact metric is not evaluated. Due to differences in speech rate, human narration exhibits natural variations and pauses compared to the TTS used in video creation, the WCD metric is not computed for human outputs. Finally, because some videos either lack subtitles or contain only partial subtitles, and because subtitle extraction is subject to recognition accuracy issues, the SSA metric is not evaluated for human outputs.

\noindent\textbf{VC-LLM(GPT-4o) VS Human.} 
As shown in Table~\ref{tab:metrics}, when evaluated on metrics such as VSC, Coh, and Logic, VC-LLM(GPT-4o) outperforms human-generated outputs. In particular, the VSC score for GPT-4o ($1.8532$) exceeds that of human outputs ($1.6276$), and similar improvements are observed in Coh ($1.9283$ vs. $1.5590$) and Logic ($1.9820$ vs. $1.8663$). Note that SRA, Fact, WCD, and SSA are not evaluated for human outputs as explained earlier.

\noindent\textbf{VC-LLM(Fine-tuned Model) VS VC-LLM(GPT-4o).}
The fine-tuned version of VC-LLM demonstrates marked improvements over the GPT-4o-based model. Most notably, the SRA increases substantially from $0.0281$ (GPT-4o) to $0.1098$ in the fine-tuned model, indicating enhanced narrative logic. Additionally, the fine-tuned model achieves a significant reduction in WCD, lowering the error from $7.2442$ to $1.1285$, and exhibits better SSA, improving from $0.3334$ to $0.9031$. Although the VSC, Fact, Coh, and Logic scores remain largely comparable, these improvements collectively suggest that fine-tuning on a curated dataset effectively enhances the overall performance of the framework.



\subsection{Ablations}

\noindent\textbf{Clip representation}
\noindent As shown in Table~\ref{tab:spatial_input},
adding spatial input improves SRA ($0.1076\rightarrow 0.1098$) and VSC ($1.8055\rightarrow 1.8069$), enhancing video sequencing and visual-script alignment. Fact increases ($1.9573\rightarrow 1.9684$), indicating better factual consistency.

\noindent\textbf{Supplement the conditions during training.}
As shown in Table~\ref{tab:condition},
adding supplementary information during SFT significantly improves factual consistency ($1.8395\rightarrow 1.9427$). However, it leads to an increase in WCD, indicating weaker control over script length. Other metrics, such as VSC, Coh, and Logic, also show slight improvements, suggesting better alignment between visual content and script, and enhanced script contextual coherence.


\noindent\textbf{Continued pretraining.}
As shown in Table~\ref{tab:continued_pretraining},
continued pre-training significantly improves SRA, indicating better narrative logic. It also enhances factual consistency (Fact$\uparrow$) and maintains high coherence (Coh) and logical consistency (Logic). Additionally, WCD decreases, suggesting stronger control over script length.

\noindent\textbf{Max frames.}
As shown in Table~\ref{tab:max_frames},
increasing the maximum frames from 1 to 5 improves SRA, Fact, and VSC scores, indicating better video-script alignment and factual consistency.

\noindent\textbf{Base model.}
As shown in Table~\ref{tab:model_comparison},
IXC2.5\citep{internlmxcomposer2_5} outperforms IXC1.0\citep{internlmxcomposer} across all script-related metrics, showing improvements in VSC, Fact, Coh, and Logic. Additionally, SSA increases, indicating better subtitle segmentation. However, WCD also increases, suggesting weaker control over script length.
Overall, the base model has a significant impact on VC-LLM, and replacing it with a better base model can enhance VC-LLM's performance.

\subsection{Online Performance}
\begin{table}[htbp]
\centering
\caption{Online A/B results (relative improvement).}
\label{tab:online_ab_test}
\begin{tabular}{c|c}
\toprule 
Group
& \makecell{adoption rate} \\
\midrule
\makecell{Baseline Group}
& - \\
\makecell{Experimental Group}
& $\textbf{+25.88\%}$ \\
\bottomrule 
\end{tabular}
\vspace{2pt}
\end{table}
\noindent A/B test has been conducted on an online platform. 
An advertiser is randomly shown with $10$ videos from two different strategies each time and can click to adopt arbitrary number of videos. 
The baseline group represents the prevailing online methodology of that period. 
It first generates a script based on the product information, then segments the script by punctuation, and finally performs matching using a fine-tuned CLIP\citep{radford2021learning} embedding model.
Since the performance of the baseline group involves trade secrets and cannot be given, we instead give the relative improvement.
As shown in Table~\ref{tab:online_ab_test}, the experimental group achieves $25.88\%$ improvement in adoption rate.
Online cases are shown in Fig.~\ref{fig:showcases}. As seen, our VC-LLM is capable of creating attractive video contents. 
\section{Conclusion}
We introduced VC-LLM, a framework leveraging Large Language Models to automatically generate short-form advertisement videos from raw footage. Experiments show that VC-LLM based on GPT-4o produces videos comparable to human-created ones, while fine-tuning on a curated dataset significantly enhances narrative logic, visual-script correlation, script quality, and subtitle quality. These results demonstrate the potential of VC-LLM to improve efficiency and creativity in advertisement video production.

\bibliographystyle{ACM-Reference-Format}
\bibliography{bibfile}
\end{document}